\documentclass[conference]{IEEEtran}
\usepackage{float}
\usepackage[noadjust]{cite}
\usepackage{tikz}
\usetikzlibrary{arrows.meta}
\usepackage{amsmath}
\usepackage{amssymb}
\usepackage{multirow}
\usepackage{footnote}
\usepackage{hyperref}
\makesavenoteenv{tabular}
\usepackage{threeparttable}

\usepackage[export]{adjustbox}
\usepackage[nottoc]{tocbibind}
\usepackage[ruled,vlined ]{algorithm2e}

\begin{document}
\title{End-to-End Deep Learning for Reliable Cardiac Activity Monitoring using Seismocardiograms}

\author{
\IEEEauthorblockN{
Prithvi Suresh\textsuperscript{$\alpha$}, Naveen Narayanan\textsuperscript{$\beta$}, Chakilam Vijay Pranav\textsuperscript{$\gamma$},
Vineeth Vijayaraghavan\textsuperscript{$\epsilon$}
}
\IEEEauthorblockA{
\textsuperscript{$\alpha\gamma\epsilon$}Solarillion Foundation, Chennai, India\\
\textsuperscript{$\beta$}SSN College of Engineering, Chennai, India\\
\texttt{
\{prithvisuresh\textsuperscript{$\alpha$}, vineethv\textsuperscript{$\epsilon$}\}@ieee.org}\\
\texttt{naveen17097@cse.ssn.edu.in\textsuperscript{$\beta$}
}\\
\texttt{cvpranav@gmail.com\textsuperscript{$\gamma$}}
}
}




\maketitle

\begin{abstract}



Continuous monitoring of cardiac activity is paramount to understanding the functioning of the heart in addition to identifying precursors to conditions such as Atrial Fibrillation. Through continuous cardiac monitoring, early indications of any potential disorder can be detected before the actual event, allowing timely preventive measures to be taken. Electrocardiography (ECG) is an established standard for monitoring the function of the heart for clinical and non-clinical applications, but its electrode-based implementation makes it cumbersome, especially for uninterrupted monitoring. Hence we propose SeismoNet, a Deep Convolutional Neural Network which aims to provide an end-to-end solution to robustly observe heart activity from Seismocardiogram (SCG) signals. These SCG signals are motion-based and can be acquired in an easy, user-friendly fashion. Furthermore, the use of deep learning enables the detection of R-peaks directly from SCG signals in spite of their noise-ridden morphology and obviates the need for extracting hand-crafted features. SeismoNet was modelled on the publicly available CEBS dataset and achieved a high overall Sensitivity and Positive Predictive Value of 0.98 and 0.98 respectively.




\end{abstract}


\section{Introduction}
In the last decade, the requirement for continuous, non-intrusive monitoring of cardiac activity has significantly increased. From athletes donning a network of sensors to maximize their performance 
to people at risk of post-operative complications,
 there is a dire need for continuous surveillance of the activity of the heart. Studies on cardiac disorders such as Atrial Fibrillation \cite{barauskiene2016importance} and Essential Hypertension \cite{pal2009spectral} show that a thorough analysis of the history of the patient's physiological signals can help avoid such catastrophes. This furthers the need for discreet cardiac monitoring. 

The Electrocardiogram (ECG) has served as the gold standard measurement of information pertaining to the function of the heart for more than a  century \cite{mortara2005ecg}. This information is obtained from the acquired ECG by detecting and processing R-peaks using various methods \cite{hrvrpeak,5738027} and extracting additional information such as Heart Rate Variability (HRV) indices. Despite the ECG providing accurate measurements of the cardiac activity, it suffers from a number of limitations. This renders it incapable of adapting to the surge in requirement of non-intrusive monitoring. ECG acquisition systems require proper placement of electrodes to avoid motion artifacts. This requires a healthcare professional, who might not be available on call in developing countries and rural areas. 
Additionally, noise free ECG signals require direct and clean skin contact which calls for obtrusive electrode placement. This makes the process of ECG acquisition stressful to patients at the best, or prone to infection or irritation of the skin at the worst \cite{lund1997disruption}.

\begin{figure}
    \centering
    \includegraphics[width = \columnwidth]{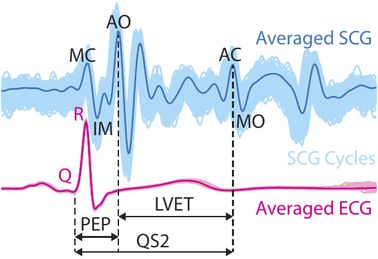}
    \caption{SCG signal vs ECG signal
    Adapted from \cite{figurepaper}}
    \label{scgvecgdiah}
\end{figure}

Another physiological signal used in cardiac monitoring is the Seismocardiogram (SCG). Seismocardiography is the monitoring of cardiac activity by measuring the acceleration produced at the sternum due to mechanical vibrations generated by myocardial motion \cite{d2019real, tadi2015seismocardiography}. Since SCG signals are extracted from the surface of the chest wall, its acquisition is less obtrusive than that of the ECG. Additionally, they can be implemented via inexpensive, miniaturized micro-electro-mechanical sensors \cite{koivisto2015automatic} that do not require an adhesive or conducting gel to be placed on the body. A combination of these factors allows for SCG to be non-intrusive, safe and viable for continuous surveillance. However, the morphology of the SCG signal is not similar to the ECG signal, as seen in Fig. \ref{scgvecgdiah}. Though the SCG is capable of yielding similar HRV indices as compared to those extracted from the ECG, it is still prone to errors since the SCG is extremely susceptible to motion artifacts \cite{7544494}.

Acknowledging the advantages and shortcomings of both these signals, we propose to derive utility from SCG signals by monitoring cardiac activity and extracting information regarding the same. In this work,  we introduce a Deep Fully Convolutional Neural Network which overcomes the lack of information in the SCG pertaining to the R-peak and robustly learns inherent patterns denoting the R-peaks' accurate position. This allows for non-invasive and safe monitoring of cardiac activity through an inexpensive, widely available sensor. The HRV indices obtained from the results of the model showcase the potential of the SCG to be used in reliable HRV monitoring.




\section{Related Work}
Over the past few decades, extensive research examining the viability of utilizing SCG signals to extract cardiac activity has been undertaken. Tadi \textit{et al.} \cite{tadi2015seismocardiography} warrants the use of SCG to obtain HRV indices by detecting the interval between consecutive Arterial Openings (AO). However, this approach required fiducial points of the ECG. In the following year, Tadi \textit{et al.} goes on to propose real-time cardiac monitoring by applying Hilbert Transform on SCG signals \cite{tadi2016real}. This approach did not require the ECG and demonstrated the viability of the SCG as a standalone signal for cardiac monitoring. Both these works prove that the HRV parameters derived from the SCG show a high correlation with the parameters extracted from ECG signals as verified by Siecinski \textit{et al.} \cite{siecinski2019comparison}. In these works, a signal processing approach is taken, where complex and computationally intensive algorithms which comprise of a cascade of signal processing operations such as filtering, noise-removal, segmentation and feature extraction for peak finding are employed.

\begin{figure}
\centering
\includegraphics[width=\columnwidth]{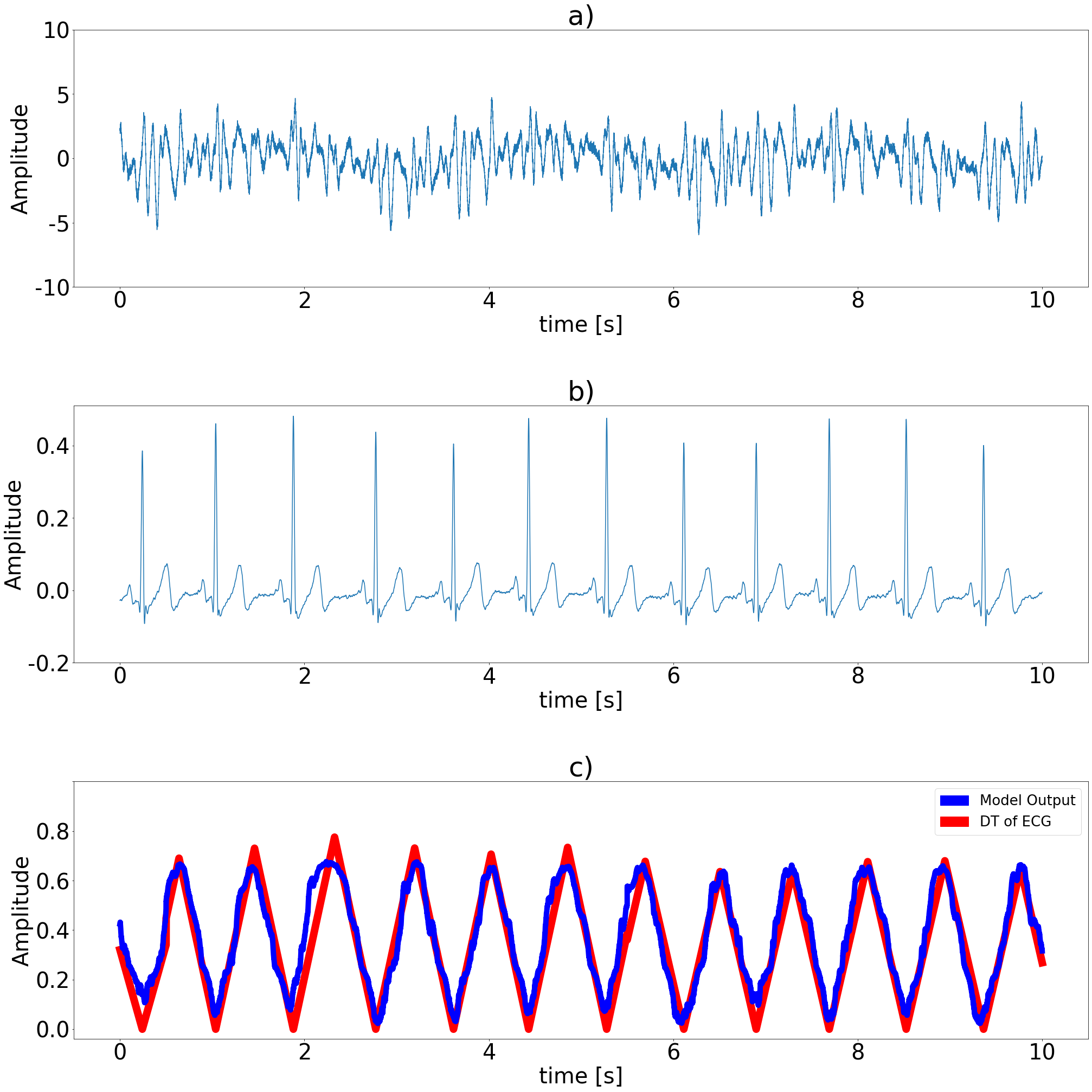}
\caption{a) SCG signal b) Corresponding ECG signal used for deriving ground truth c) Model Output vs Distance transform (DT) of ECG}
\label{fig:output}
\end{figure}

Using SCG to monitor cardiac activity with smart devices in a real-time setting has been explored by the likes of Haesher \textit{et al.} [\citenum{haescher2015study}], [\citenum{haescher2016seismotracker}] and  Hernandez \textit{et al.} \cite{hernandez2014bioglass}, \cite{hernandez2015biophone}. Similar to prior work, these make use of a barrage of signal processing algorithms and transforms. Additionally, AO peaks, which temporally occur after the R-peak in an ECG signal (refer Fig. \ref{scgvecgdiah}), are used as the fiduciary points to measure heart activity with the underlying assumption that the AO-AO intervals in an SCG signal correspond to the R-R intervals in an ECG signal. This is however not the case in a real-time setting where the morphology of an SCG signal is affected by many factors other than cardiac activity \cite{Lee_2018}  which renders them prone to noise as shown in Fig. \ref{fig:output}a. This consequentially makes it much more challenging to perform peak detection on them.

Recently, there has been acknowledgement of the virtues of application of Machine Learning (ML) and Deep Learning (DL) to analyze SCG signals and derive descriptive and predictive information from them. Yao \textit{et al.} \cite{8485348} presents a three layer artificial neural network that uses a combination of both ECG and SCG signals on a beat-by-beat basis for personalized quiescence prediction. Though this is applied specifically to the field of Coronary Computed Tomography Angiography, it portrays the successful application of ML to SCG signals. Mora \textit{et al.} \cite{articleMora} explored the application of unsupervised learning on SCG signals through the use of a variational autoencoder which was used to extract user information from their corresponding SCG waveforms. Haescher \textit{et al.} \cite{inproceedings} proposed a strategy which aims to transform SCG signals into ECG signals using a convolutional autoencoder. This was followed by the use of the Pan Tompkins algorithm \cite{pan1985real} for R-peak detection on the resultant of the transformation. The algorithm has been studied on noise ridden ECG which showed deteriorating performance \cite{liu2018performance}.\\ 
Despite a continuously rising number of devices monitoring non-medical grade signals in everyday life, there is still a dearth of reliable analysis of these signals. By leveraging end-to-end deep learning, the gap between non-clinical signal acquisition and reliable signal analysis can be bridged. We attempt to bridge this gap by proposing an end-to-end deep learning network that robustly extracts information from the prevalent SCG signals. 

\section{Problem Statement}
Although existing solutions to non-invasive cardiac monitoring through SCG signals are competent, there is still a lack of an end-to-end solution which is computationally inexpensive. Existing solutions do not take into account the morphological changes that the SCG might undergo in a real-time setting. Additionally, standalone SCG signal monitoring loses out on vital information regarding the R-peaks, thus generating inaccurate HRV indices.

Considering the aforementioned drawbacks that current solutions possess, we propose \textit{SeismoNet}, a Deep Fully Convolutional Neural Network. SeismoNet accurately detects R-peaks and hence provides robust monitoring of the SCG signal. This is achieved by training SeismoNet in an end-to-end manner to detect position of the R-peaks in spite of noise. This also dismisses the need for multiple signal processing blocks that are computationally taxing. The ease of signal acquisition of the SCG coupled with the reliability of SeismoNet facilitates user-friendly, non-invasive and continuous heart monitoring.

\begin{figure*}[!ht]
    \centering
    \includegraphics[width = \textwidth]{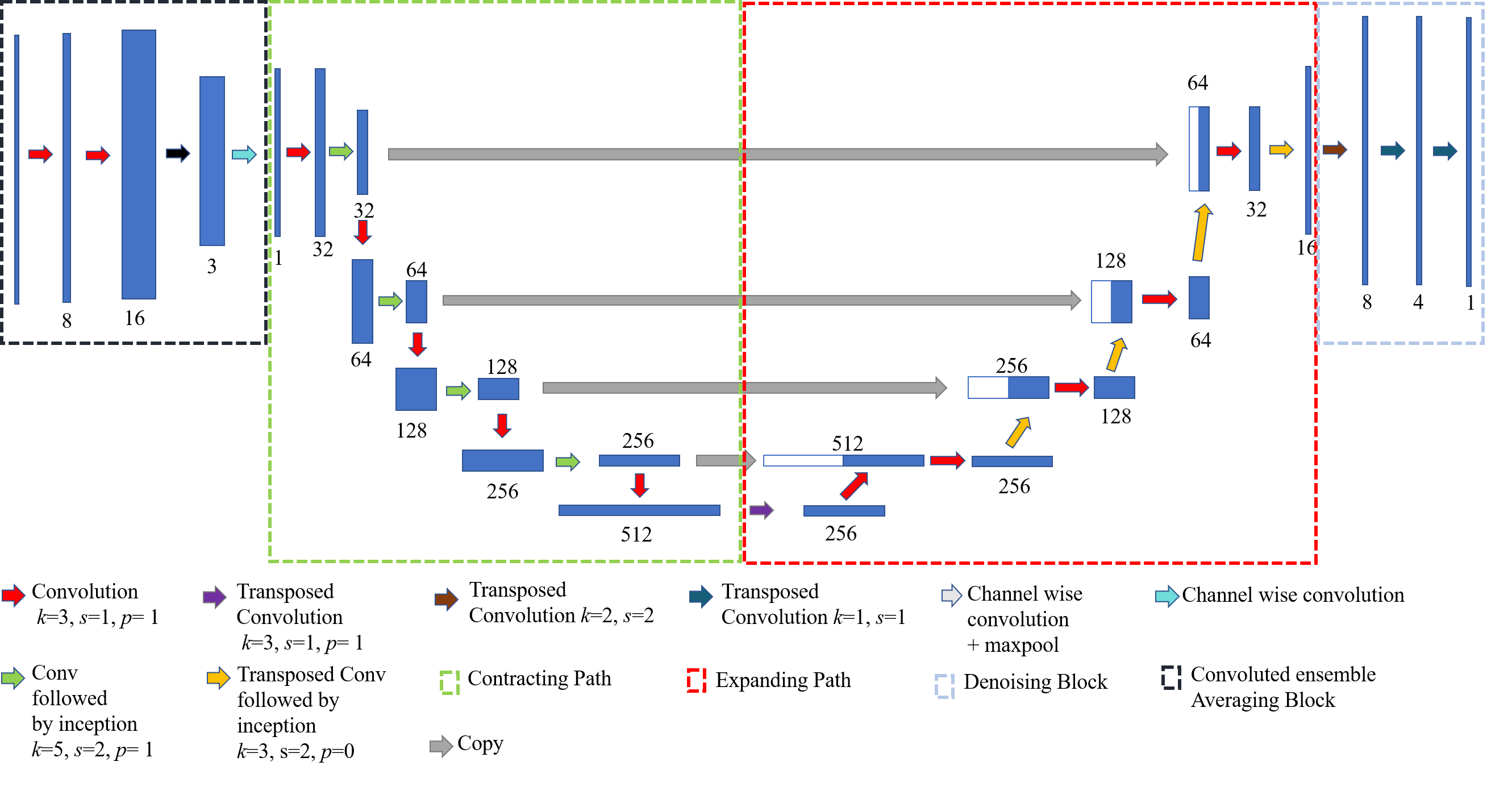}
    \caption{SeismoNet Architecture ($k$-kernel size, $s$-stride, $p$-padding)}
    \label{architecure}
\end{figure*}
\section{Methodology}
\subsection{Dataset Description and Preprocessing}
The experiments were carried out on the Combined Measurement of ECG, Breathing and Seismocardiograms (CEBS) dataset, publicly available as part of Physiobank \cite{6713413}. The dataset consists of SCG, Respiratory and ECG signals from 2 leads (leads I and II),  sampled at 5kHz. The data was collected from 20 subjects with no known ailments with a mean age of 24.4 years (SD $\pm$ 3.10). The signals were acquired via a Biopac MP36. Out of the 20 subjects only 18 were considered since subject 4 and subject 18 had discrepancies during data collection.

The protocol for data collection required subjects to lie in a supine position. Recordings were first obtained from the subject while they were in the basal state. Following this, subjects were made to listen to music while the collection procedure continued. The procedure was terminated 5 minutes after the music session ended. 

The lead I ECG ($e$) and SCG ($s$) from each subject's data were segmented into windows of length ($w$) 10 seconds with 5 seconds overlap. No filtering was performed on the SCG to best replicate real-time data reception. The resultant windows from each subject were split into 3 sections (60\%-20\%-20\%) for training ($X$), validation ($V$) and testing ($T$) respectively. 

\subsection{Signal Paradigm}
\label{prob}
We propose an architecture that transforms the SCG signal ($s^{(i)},\ s^{(i)}\in X$) into an interpretable waveform consisting of relevant information pertaining to the corresponding ECG signal ($e^{(i)}$). This allows for extraction of heart rate indices through elementary algorithms that have almost no latency. Thus, a  signal-to-signal model is used whose architecture is elaborated in the following section.

SeismoNet maps the SCG signal to a transformed ECG signal ($t^{(i)}_{gt}$), wherein $t^{(i)}_{gt} = DT(e^{(i)})$ is the Distance Transform of the ECG signal. 
Traditionally used in Computer Vision applications \cite{gavrila1998multi, felzenszwalb2012distance}, Distance Transforms are used to map the distance of a pixel to another pixel of a specific type (such as a boundary pixel). Accordingly, the Distance Transform of an ECG signal is defined as the distance between each sample in the signal to the closest R-peak annotation. This results in a transformed signal as shown in Fig. \ref{fig:output}c. The valleys of this transformed signal signify the position of the R-peak. 

Thus, it is necessary to optimize the weights of the model to produce a signal that resembles $t^{(i)}_{gt}$. This is carried out by minimizing the $Smooth_{L1}$ loss between the prediction of the model ($t^{(i)}_{pred}$) and the ground truth ($t^{(i)}_{gt}$). The loss function $\mathcal{L}$ is given by
\begin{equation}
\label{loss}
    \mathcal{L}= \sum_{i=1}^m Smooth_{L1}(t^{(i)}_{pred} - t^{(i)}_{gt})
\end{equation}
\begin{equation}
    Smooth_{L1}(x) =\begin{cases}
  0.5x^2, & \text{if }|x|<1\\    
  |x| - 0.5, & \text{otherwise}    
\end{cases}
\end{equation}

\subsection{SeismoNet Architecture}
The architecture of SeismoNet is an extension of the U-net framework proposed by Ronneberger \textit{et al.} \cite{ronneberger2015u}. The U-net repeatedly contracts (contracting path) the input, followed by repeated expansion (expanding path) until the same dimensions of the input is obtained, akin to a standard encoder-decoder. U-net uses skip connections that connect the activations learnt by the deeper layers to the shallower layers.

\begin{figure}
    \centering
    \includegraphics[width =\columnwidth]{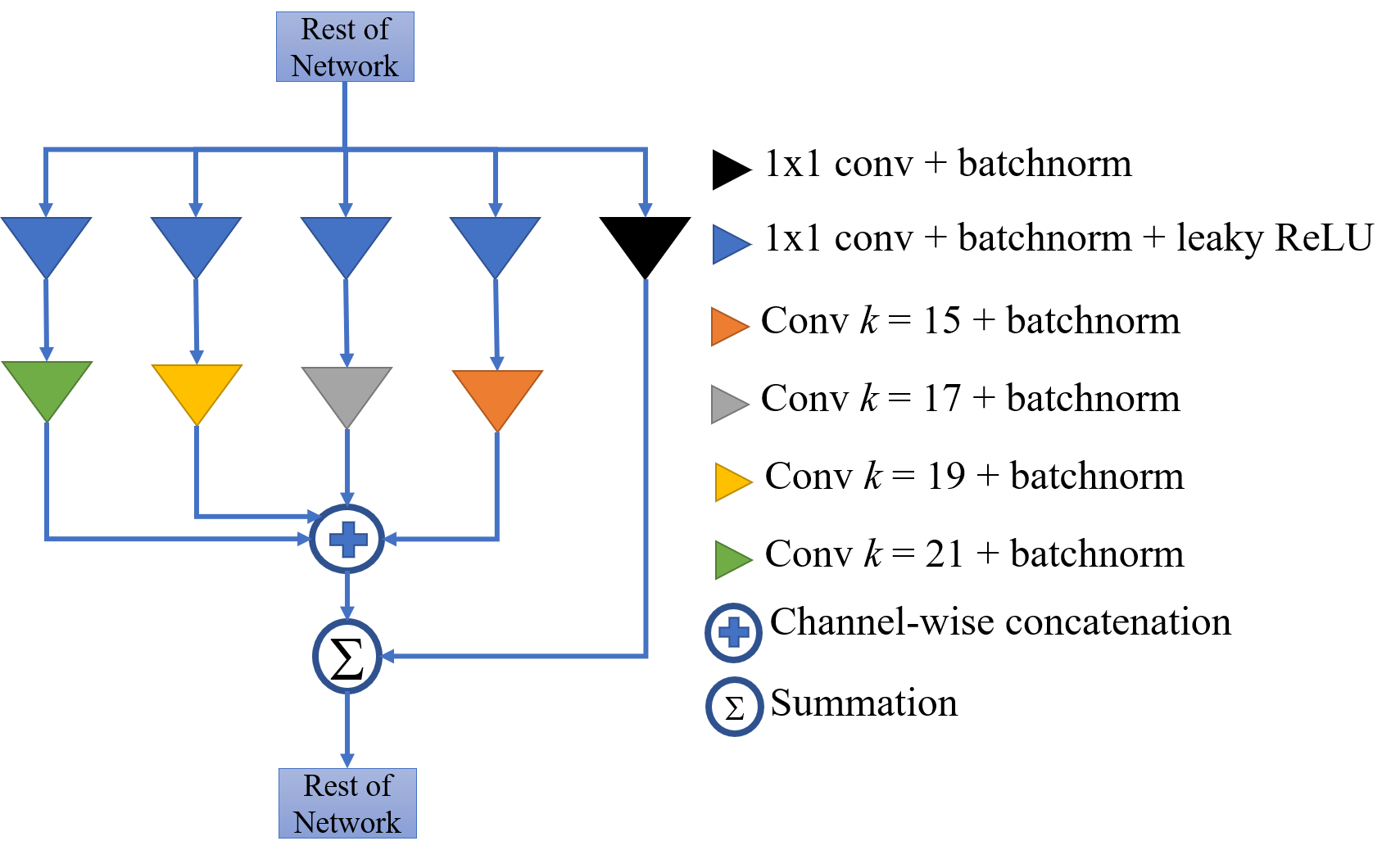}
    \caption{Inception Residual Block}
    \label{incres}
\end{figure}

The architecture proposed is broken down into blocks depending on their purpose. There are a total of $2N + 2$ blocks in the network made up of $N$ Contracting Convolutional Blocks (CCB), $N$ Expanding Convolutional Blocks (ECB), 1 Convolutional Ensemble Averaging Block and 1 Denoising Block. The Inception-Residual Block has been excluded while counting since it features repeatedly in other blocks.  The purpose and structure of each block is as follows:
  
\begin{enumerate}
     \item \textbf{Inception-Residual Block}: The Inception-Residual block consists of multiple convolutions\footnote{All convolutions refer to one dimensional convolutions unless specified.} with different kernel sizes performed in parallel as shown in Fig. \ref{incres}. Following this, channel wise concatenation of these representations is done. Furthermore, a residual connection is introduced to alleviate vanishing gradients. 
    
    \item \textbf{Contracting Convolutional Block (CCB)}: Each Contracting Convolutional Block ($C_n^c$, where $n\in [1,N]$ and $c$ is number of input channels) consists of a padded convolutional layer (kernel size = $k_P$) that doubles the number of channels. This is succeeded by batch normalization and application of Leaky ReLU activation. This is then followed by strided convolution (kernel size = $k_S$) for learnt-downsampling before being fed to an Inception-Residual Block to produce the block's output feature map $f_{C_{n+1}^{c'}}$, where $c' = 2c$. 
    \item \textbf{Expanding Convolutional Block (ECB)}: 
    Each Expanding Convolutional Block ($E_n^c$) has a padded convolution layer of kernel size $k_P$ that halves the number of channels, followed by batch normalization and leaky ReLU activation. Strided transposed convolution of kernel size $k_{S^T}$ further reduces the number of channels by half after which the Inception-Residual Block follows. Each input feature map to the Expanding Convolutional Block consists of $f_{E_{n-1}^{c}} + f_{C_{m}^{c}}$, where $m = N-(n-1)$. This represents concatenation of the output feature map of the previous ECB with the output feature map of the corresponding CCB in the contracting path. This allows for projection of features learnt at different stages of the contracting path onto the final feature map. The output feature map of this block is given by $f_{E_{n}^{c'}}$, where $c' = c/4$ 
    
    \item \textbf{Convolutional Ensemble Averaging Block}:
    This block prepares the input signal to be fed to the contracting path. Ensemble Averaging is a statistical method used to improve the Signal to Noise Ratio by averaging out multiple versions of the same signal \cite{ensembleavg}. Retaining this intuition, the number of channels of the input signal is increased by passing it through a convolutional layer. Following this, convolution across the channels is performed which yields a surrogate signal that is jitter free with more prominent peaks, allowing for faster convergence during training.

    \item \textbf{Denoising Block}: This block serves two purposes. \\1) It allows for a better representation of the output from the U-net, ultimately producing a less distorted signal. 2) It allows for learnt-upsampling, since the output of the U-net does not match the dimensions of the input signal exactly. 
    
\end{enumerate}
The input signal of length $w$ is fed to the Convolutional Ensemble Averaging Block, which generates a feature map to be passed onto the first CCB in the contracting path. The feature map is passed through multiple CCBs until $n_{CCB} = N$, in which case the contracting path terminates and the expanding path begins. The input to the first ECB in the expanding path is the output feature map of the final CCB in the contracting path $f_{N}^{M}$, where $M = 2^{N-1}\times c_i$ and $c_i$ is the number of output channels of the first CCB. Similar to the contracting path, the feature map is passed through multiple ECBs until $n_{ECB} = N$. The final output signal is generated by passing the output of the final ECB to the denoising block. The final output signal has the same dimensions as the input signal.

\begin{figure*}[!ht]
\centerline{\includegraphics[width=\textwidth]{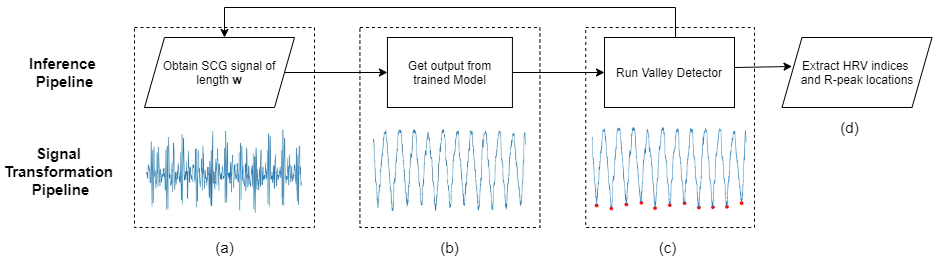}}
\caption{Inference Pipeline}
\label{pipe}
\end{figure*}
\section{Experiments and Results}
\subsection{Implementation}
For our experiments, $N=5$ was chosen which results in a total of 12 blocks. Within each CCB and ECB, we select $k_P$ to be 3 and both $k_S$ and $k_{S^T}$ to be 5. The number of output channels of the first CCB ($c_i$) is chosen to be 32. \\
The weights of the model were initialized in accordance with the Xavier uniform initializer \cite{glorot2010understanding}. Stochastic Gradient Descent with an initial learning rate of 0.001 was used to optimize the weights of the model by minimizing the loss given by Equation \ref{loss}. SeismoNet was trained for 300 epochs with a learning rate scheduler which reduced the learning rate by a factor of 10 after every 100 epochs. These constants were empirically decided after extensive experimentation.
\subsection{Inference}
As stated in section \ref{prob} the output waveform of the model yields information pertaining to the ECG signal. To obtain the output waveform, an SCG signal of window length $w$ (Fig. \ref{pipe}a) is extracted and passed through SeismoNet. SeismoNet generates an output waveform of the same length $w$ as shown in Fig. \ref{pipe}b. The valleys of the output waveform are detected to identify the R-peaks. The following subsections evaluate the different aspects of the model's performance. 

\begin{table}

\centering\caption{R-peak Detection Performance across all users}
\renewcommand{\arraystretch}{1.3}
\resizebox{\columnwidth}{!}{
\begin{tabular}{|c|c|c|c|c|c|c|c|}
 \hline
  \textbf{Subject} & \begin{tabular}[c]{@{}l@{}}\textbf{Total} \\ \textbf{Detected}\\ \textbf{Peaks}\end{tabular} & \begin{tabular}[c]{@{}l@{}}\textbf{Total} \\ \textbf{Actual}\\ \textbf{Peaks}\end{tabular} & \textbf{TP} & \textbf{FP} & \textbf{FN} & \textbf{Se} & \textbf{PPV} \\
   
 \hline
 1 & 319 & 323 & 304 & 15 & 19 & 0.94 & 0.95\\
 2 & 317 & 316 & 309 & 8 & 7 & 0.98 & 0.97\\
 3 & 367 & 368 & 365 & 2 & 3 & 0.99 & 0.99\\
 5 & 373 &375 & 370 & 3 & 5 & 0.99 & 0.99\\
 6 & 323 & 323 & 318 & 5 & 5 & 0.98 & 0.98\\
 7 & 269 & 270 & 267 & 2 & 3 & 0.99 & 0.99\\
 8 & 515 & 516 & 510 & 5 & 6 & 0.99 & 0.99\\
 9 & 321 &321 & 318 & 3 & 3 & 0.99 & 0.99\\
 10 & 311 & 310 & 306 & 5 & 4 & 0.99 & 0.98\\
 11 & 368 & 368 & 361 & 7 & 7 & 0.98 & 0.98\\
 12 & 443 & 440 & 432 & 11 & 8 & 0.98 & 0.98\\
 13 & 378 &378 & 374 & 4 & 4 & 0.99 & 0.99\\
 14 & 351& 350 & 344 & 7 & 6 & 0.98 & 0.98\\
 15 & 337 & 337 & 334 & 3 & 3 & 0.99 & 0.99\\
 16 & 390 &391 & 387 & 3 & 4 & 0.99 & 0.99\\
 17 & 356 & 353 & 345 & 11 & 8 & 0.98 & 0.97\\
 19 & 348 & 344 & 336 & 12 & 8 & 0.98 & 0.97\\
 20& 352 & 354 & 343 & 9 & 11 & 0.97& 0.97\\
\hline
 Total & 6438 &6437& 6323 & 115 & 114 & \textbf{0.98} & \textbf{0.98}  \\
 \hline
\end{tabular}    
\label{table:1}
}

\end{table}

\subsection{R-peak Detection}
\subsubsection{Metrics}
Sensitivity ($Se$) and Positive Predictive Value ($PPV$) were used to assess the model's efficacy in making accurate R-peak predictions whilst maintaining an immunity to falsely detected and missed peaks. If SeismoNet's detection falls within 90 ms \cite{siecinski2019comparison} of the actual peak, it is classified as a True Positive (TP), otherwise it is classified as a False Positive (FP). Actual peaks which have been omitted from the predictions are classified as False Negatives (FN). Sensitivity and Positive Predictive Value are used to quantify SeismoNet's R-peak detection performance. The Sensitivity and Positive Predictive Value are defined using equations (\ref{eq:Se}) and (\ref{eq:PPV}) respectively.

\begin{equation}
 \label{eq:Se}
Se=\frac{TP}{TP+FN} 
\end{equation}

\begin{equation}
 \label{eq:PPV} 
PPV=\frac{TP}{TP+FP} \end{equation}
\subsubsection{Performance}
Table \ref{table:1} shows the performance of the model on the Test Set across all users. The total number of R-peaks in the test set was 6438 out of which 6323 were correctly detected. The sensitivity ranges from 0.94 in the case of  subject 1, all the way to 0.99 for multiple subjects. Similarly, the lowest $PPV$ was observed to be 0.95 for Subject 1 and ranged between 0.97-0.99 for all the other subjects. 
On further analysis of subject 1, it was observed that the SCG signal was corrupted with substantial noise. Inspite of this, SeismoNet performed significantly better even in the presence of a strict tolerance of 90 ms.

\begin{table}
\renewcommand{\arraystretch}{1.3}
\centering 
\caption{HRV Parameters}
\label{HRVparams}
\begin{tabular}{|c|c|}
\hline
\textbf{HRV Parameter} & \textbf{Definition}                                                                                                                                               \\ \hline
Mean NN (ms)           & Arithmetic mean of all interbeat intervals                                                                                                                        \\ \hline
SDNN (ms)              & Standard Deviation of all interbeat intervals                                                                                                                     \\ \hline
RMSSD (ms)             & \begin{tabular}[c]{@{}c@{}}Root mean square of successive differences \\ between heart beat intervals\end{tabular}                                                \\ \hline
pNN50                  & \begin{tabular}[c]{@{}c@{}}Ratio of total number of successive NN intervals \\ whose difference exceeds 50 ms to the total \\ number of NN intervals\end{tabular} \\ \hline
\end{tabular}
\end{table}            

\subsection{HRV Indices Extraction}
The most clinically significant HRV parameters in the time domain were extracted by analysing the R-peaks detected by the model. This was done in order to assess their serviceability, by comparing them with the actual HRV indices obtained from the ECG signal. The parameters considered are shown in Table \ref{HRVparams}. 


\subsubsection{HRV indices Performance}
\begin{figure}
\renewcommand{\arraystretch}{1.3}
\begin{tabular}{cc}

\centering
  \includegraphics[width=40mm]{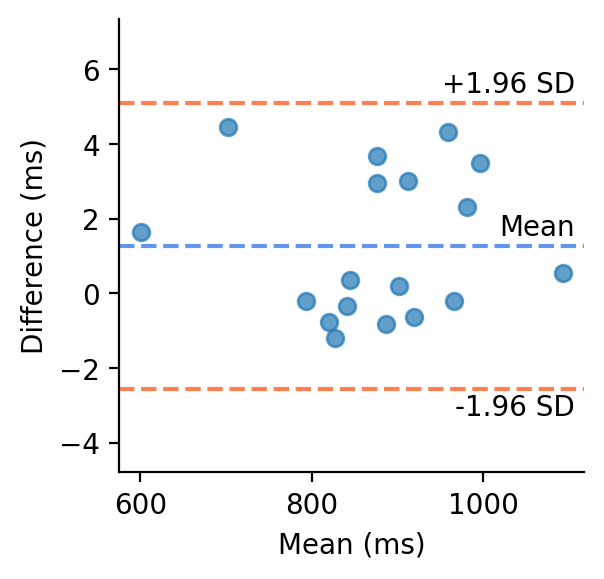} &   \includegraphics[width=40mm]{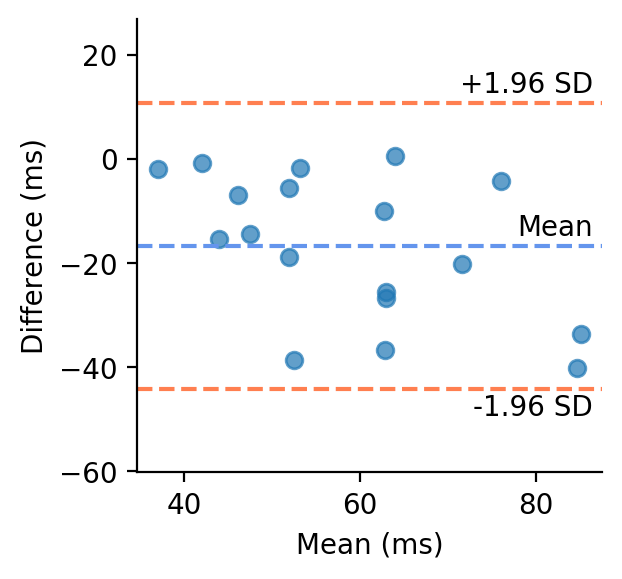} \\
(a) Mean NN  & (b) SDNN (54.86 ms)\\
LoA range = 7.65 ms& LoA range = 54.86 ms \\[6pt]
 \includegraphics[width=40mm]{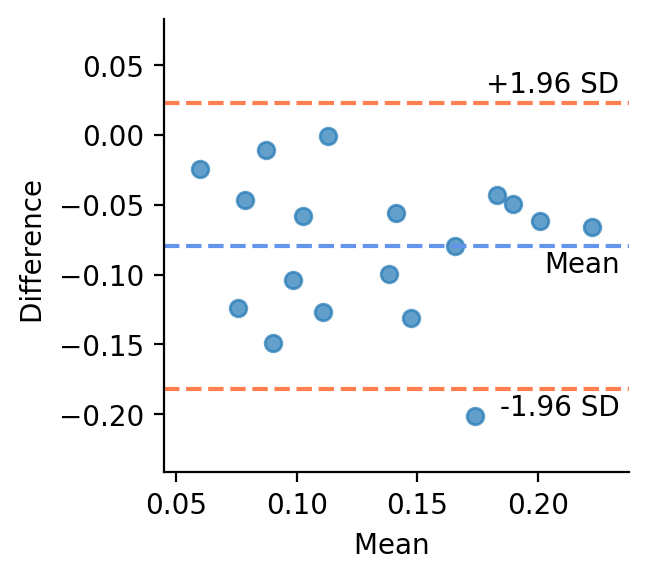} &   \includegraphics[width=40mm]{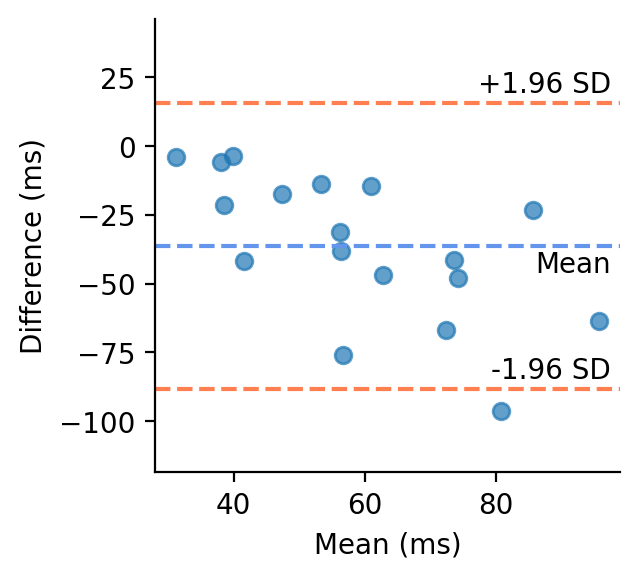} \\
(c) pNN50 (0.20) & (d) RMSSD (103.78 ms)\\
 LoA range = 0.20 & LoA range = 103.78 ms\\[6pt]
\end{tabular}
\caption{Bland-Altman Plots of subject-wise HRV Indices with their corresponding range of LoA}
\label{blandaltmanplots}
\end{figure}

To understand the agreement between the HRV indices extracted from the R-peaks generated from the model (SCG-indices) and from the R-peaks annotated on the ECG (ECG-indices), a Bland-Altman analysis is carried out. In the Bland-Altman plot, the error between two corresponding quantities is plotted against their mean, which in turn is used to systematically identify outliers in the data. In our case, the two quantities were the subject-wise SCG-indices and ECG-indices. The Limits of Agreement (LoA) were chosen to be $\pm1.96\ SD$ which corresponds to a confidence interval of 95\%. As seen from each plot in Fig. \ref{blandaltmanplots}, the range of the LoA is sufficiently small indicating high agreement of the SCG-indices and ECG-indices. 100\% of the points lie within the LoA in the plots corresponding to the mean NN and SDNN as seen in Figures \ref{blandaltmanplots}a and \ref{blandaltmanplots}b respectively. This implies a high correlation between the two quantities. From Fig. \ref{blandaltmanplots} it can also be seen that there is only one outlier in the plots of pNN50 and RMSSD. These points correspond to subject 1.

The absolute values of the SCG-indices and ECG-indices across all users are shown in Table \ref{hrvindices}. The difference between the indices is negligible thus validating the performance of SeismoNet as a tool to obtain accurate HRV as well.

\begin{table}
\renewcommand{\arraystretch}{1.3}

\centering
\caption{HRV Indices derived from ECG and SCG}
\begin{tabular}{|l|r|r|r|r|}
\hline
\multicolumn{1}{|c|}{\multirow{2}{*}{\textbf{HRV Indices}}} & \multicolumn{2}{c|}{\textbf{SCG}}                            & \multicolumn{2}{c|}{\textbf{ECG}}                            \\ \cline{2-5} 
\multicolumn{1}{|c|}{}                             & \multicolumn{1}{c|}{\textbf{Mean}} & \multicolumn{1}{c|}{\textbf{SD}} & \multicolumn{1}{c|}{\textbf{Mean}} & \multicolumn{1}{c|}{\textbf{SD}} \\ \hline
Mean NN (ms)                                               & 876.6189                  & 108.5728                & 877.8914                  & 108.6028                \\ \hline

SDNN (ms)                                              & 67.2341                   & 18.0767                 & 50.5093                   & 11.7741                 \\ \hline
RMSSD (ms)                                              & 77.3436                   & 27.8062                 & 41.0370                 & 13.7633                 \\ \hline
pNN50                                              & 0.1720                   & 0.0552                 & 0.0925                   & 0.0516                  \\ \hline

\end{tabular}
\label{hrvindices}
\end{table}





\section{Conclusion}

A Deep Fully Convolutional Neural Network was designed for creating an end-to-end solution for continuously examining cardiac activity using SCG signals, in a manner that is convenient and robust. Our model, SeismoNet, was able to detect R-peaks from SCG signals from the CEBS dataset, with a total Sensitivity of 0.98 and Positive Predictive Value of 0.98. On further analysis using Bland Altman plots, it was found that the HRV parameters derived from the SCG signals and from the ECG signals showed high correlation, thereby validating their agreement and interchangeability. Therefore, SeismoNet facilitates continuous monitoring of cardiac motions with exceptional accuracy while avoiding complex signal processing techniques and also keeping in mind the simplicity of such a system in real-world scenarios.





\bibliographystyle{IEEEtran}
\bibliography{citations}
\end{document}